\documentclass[conference]{IEEEtran}
\IEEEoverridecommandlockouts

\usepackage[utf8]{inputenc} 
\usepackage[T1]{fontenc}    
\usepackage{hyperref}       
\usepackage{url}            
\usepackage{booktabs}       
\usepackage{amsmath,amssymb,amsfonts}
\usepackage{nicefrac}       
\usepackage{microtype}      
\usepackage{graphics} 
\usepackage{epsfig} 
\usepackage{multirow}

\usepackage{cite}

\usepackage{algorithm}
\usepackage[noend]{algpseudocode}

\usepackage{bm}
\usepackage{graphicx}
\usepackage{textcomp}
\usepackage{xcolor}
\usepackage{comment}
\usepackage{verbatim}

\def\BibTeX{{\rm B\kern-.05em{\sc i\kern-.025em b}\kern-.08em
    T\kern-.1667em\lower.7ex\hbox{E}\kern-.125emX}}

\definecolor{mypink}{RGB}{219, 48, 122}

\begin{document}

\title{Detection of Real-world Driving-induced \\Affective State Using Physiological Signals \\and  Multi-view Multi-task Machine Learning
}

\author{\IEEEauthorblockN{Daniel Lopez-Martinez}
\IEEEauthorblockA{\textit{Harvard-MIT Health Sciences and Technology and MIT Media Lab} \\
\textit{Massachusetts Institute of Technology}\\
Cambridge, USA \\
dlmocdm@mit.edu}
\and
\IEEEauthorblockN{Neska El-Haouij and Rosalind Picard}
\IEEEauthorblockA{\textit{MIT Media Lab} \\
\textit{Massachusetts Institute of Technology}\\
Cambridge, USA \\
$\{$neska, picard$\}$@media.mit.edu}
}


\maketitle

\begin{abstract}
Affective states have a critical role in driving performance and safety. They can degrade driver situation awareness and negatively impact cognitive processes, severely diminishing road safety.
Therefore, detecting and assessing  drivers' affective states is crucial in order to help improve the driving experience, and increase safety, comfort and well-being. Recent advances in affective computing have enabled the detection of such states. This may lead to empathic automotive user interfaces that account for the driver's emotional state and influence the driver in order to improve safety.  In this work, we propose a multi-view multi-task machine learning method for the detection of driver's affective states using physiological signals. The proposed approach is able to account for inter-drive variability in physiological responses while enabling interpretability of the learned models, a factor that is especially important in systems deployed in the real world. We evaluate the models on three different datasets containing real-world driving experiences. Our results indicate that accounting for drive-specific differences significantly improves model performance.

\end{abstract}
 

\section{Introduction}
Affective automotive user interfaces are being developed by most large automotive companies, recognizing  that the interaction with the driver can impact driver safety not only via distraction, but also via mismatched affective state.  For example, Nass and team \cite{Nass2005,Johnsson2005} showed that if the car navigation system's tone of voice is kept the same, then the tone that works best when a driver is in a happy state is the tone that works worst when the driver is in a mildly upset state, and vice-versa. 
Using the wrong tone of voice thus results in a higher rate of accidents. Consequently, an interface that has more human-like social-emotional intelligence would know how to change its tone of voice to optimize safety. However, the decision of what type of vocal tone is safest requires understanding the driver's affective state at the moment before speaking. Hence, the interface needs continuous, real-time monitoring and decision-making about affect {\it before} it speaks.    

A system with social-emotional intelligence may also help drivers in other ways. For example, it might sense a context or physiological change consistent with likely stress, frustration, or anger in a driver, or with heightened arousal and exuberance. Human affective state is complex and continuously changing depending on internal and external triggers. Both positive and negative emotions can alter any task performance. In the context of daily driving, excessive distress, for instance, can negatively affect overall well-being  as well as road safety. Therefore, an interface that subtly assists the driver in coping with  environmental demands and maintaining an adequate affective state will become able to provide significant value beyond simply providing a more pleasant driving experience.

Researchers, automakers and automotive manufacturers are using recent affective computing innovations to better understand the driver's optimal affective state for achieving safe and comfortable driving experiences.
With  recent advances in non-intrusive and wearable sensors, continuous information about drivers' affective states can be objectively assessed. The detection of such states can be used in the design of closed-loop interventions. Several studies have evaluated driver's state prediction using physiological signals, with features extracted from heart rate (HR), electrodermal activity (EDA), breathing rate (BR), electromyogram (EMG) and electroencephalogram (EEG), showing promising accuracy for driver's affective state recognition \cite{Healey2005,Rigas2008,Lee2012,Singh2013,Vincente2016,Chen2017,Saeed2017,Kim2018, ElHaouijSMA}.

Unlike experiences achieved using driving simulators where the environment is reproducible and controlled, each real-world driving context contains unexpected and changing events. In addition, it is time- and cost-consuming to conduct real-road driving experiments. For that, it is a challenge to collect and to identify the driver's true underlying affective state. Few physiological databases, collected during real-world driving experiences, are publicly accessible.

In this work, we take three publicly available datasets collected from real-world driving and use them to develop and investigate a machine learning method for the inference of affective state.
Specifically, we propose a multi-view multi-task machine learning approach to affect recognition during real-world driving based on two physiological signals that can be unobtrusively acquired: electrodermal activity (EDA) and heart rate (HR). The deployment of machine learning systems in complex applications where safety is critical requires systems that are optimized for performance but also offer interpretability of their reasoning, so that we may verify whether that reasoning is sound \cite{doshiKim2017}. Therefore, we show how our method is able to account for drive-specific differences in physiological responses, resulting in improved performance, while enabling interpretability of the learned model. Across the three datasets, we show that features extracted from physiological signals, coupled with multi-view multi-task machine learning, can provide a significantly accurate prediction of driver's state and can cluster the driving performance into different physiological profiles.

This work makes three main contributions: (1) We present an interpretable machine learning model for affective state recognition during driving; (2) We evaluate the model on three different datasets containing real-world physiological driving data; (3) We show that accounting for inter-drive variability using a multi-task approach significantly improves affective state recognition performance, with respect to single-task models. The resulting approach is thus able to be implemented in real-world automotive user interfaces to provide greater social-emotional intelligence during continuous driving interactions.

\section{Datasets} \label{sec:data}
In order to evaluate our approach to driver's affective state recognition, we used three public databases containing physiological signals recorded from different real-world driving experiences: MIT drivedb, HciLab, and AffectiveROAD. 
\subsection{MIT drivedb}

The MIT drivedb \cite{Healey2005} dataset contains data from  $20$ miles of driving in the greater Boston area, USA. The driving protocol was as follows. Before and after driving, participants underwent a $15$-minute rest period, from which data were collected. Then, participants drove through pre-assigned routes that alternated  between city and highway driving. From these experiences, the following physiological signals were recorded: EDA measured in two placements (hand and foot), ECG, EMG and BR. Videos recording the car's inside and outside scenes were used by experts to compute the \textit{distress} level in each frame. This computed score was found correlated with the road type \cite{Healey2005}. A part of the database, is publicly available on PhysioNet \cite{Goldberger2000} and on the MIT Media Lab website \cite{Databases}. More precisely, datasets related to $17$ out of $24$ drives are released publicly. They mainly contain physiological data recorded during each drive, in addition to a signal reflecting the onset and offset of each experiment condition (rest period, city, or highway driving). The latter signal is used to derive the distress label based on the road type. In fact, the stress metric proposed in the work of Healey and Picard \cite{Healey2005} validated the assumption of low, medium and high stress levels during the rest, highway and city driving, respectively.         

\subsection{HciLab data}
The HciLab database \cite{Schneegass2013} consists of data collected during drives performed in Stuttgart, Germany. It contains signals recorded during $10$ drives performed by $10$ participants, who drove through different routes including both highway and freeway, as well as city driving with two different speed limits. Each participant used his/her own car in order to avoid the workload increase characteristic of the initial adaptation to a new vehicle. The recorded physiological data are EDA, ECG and body temperature. Brightness, acceleration and videos recording the inside and outside driving scenes were collected. The authors of the database propose a \textit{workload} metric with each drive. This metric was annotated post-experience by the driver, while watching videos of his/her driving test. The score of the video rating is recorded between $0$ (no workload) to $128$ (maximum workload).

\subsection{AffectiveROAD data}
AffectiveROAD \cite{ElHaouij2018} is a publicly available dataset \cite{Databases} collected in Tunisia following the driving protocol of the MIT drivedb dataset \cite{Healey2005}. In total, it contains $13$ drives performed by $10$ drivers. The following physiological signals were recorded: EDA measured on both wrists, HR, BR and skin temperature. In addition, data related to GPS, videos (filming inside and outside car scenes) and in-car temperature, humidity and sound level were collected. 
A \textit{stress} metric is proposed in this database, which was constructed in real-time by an observer seating in the rear seat of the car, who annotated the overall perceived stress for the entire drive from low (score=$0$) to high (score=$1$) using a slider . For each drive, this subjective stress metric was validated after the driving experiment by the driver, who was shown the video recordings and asked to correct and validate the continuous score.

\section{Methodology}

We describe in this section the features and labels extracted from the different datasets. In addition, the machine learning approach, consisting of spectral clustering and multi-task machine learning, is detailed.

\subsection{Feature and label extraction} \label{sec:featureslabels}
Feature extraction was performed on sliding windows extracted from the electrodermal activity (EDA) and heart rate (HR) data. 
The EDA, defined as a signal reflecting electrical activity measured from the skin, has been used in several studies related to human affect recognition, with many studies finding EDA to be correlated with human stress levels \cite{Boucsein2012}. The EDA and HR were selected since they are relevant signals in stress recognition, especially for driving task performance \cite{Healey2005,Singh2013,Vincente2016,ElHaouijSMA}. In addition, those physiological signals are selected since they can be found in all three databases.

The EDA (captured on the left side of the driver) and the HR signals were extracted from all three databases. The EDA signal was passed through a low pass filter to eliminate high frequency noise, similar to Healey \cite{Healey2000}. Subsequently, EDA and HR are normalized. The normalization technique used to compensate the individual differences was based on the widely used min-max range normalization. From these signals, following previous work \cite{Saeed2017}, we extracted 30-second windows with an overlap of $50\%$ (15 seconds). For each window, statistical features are computed for: mean, standard deviation, minimum, and maximum. For the EDA, the kurtosis and skewness were computed in addition to three peaks features: total number of peaks, their total amplitude and duration. For peaks (startle) detection, we used the same approach as used in Healey's thesis \cite{Healey2000}. For the HR, the root mean square of successive differences is computed. 

The drivers' affective state label, corresponding to each segment, was extracted differently for each database. For drivedb, we were able to consider three levels of stress based on the driving period extracted the annotation file. The high stress level (labelled \texttt{H}) is assumed to be evoked by city driving, medium stress level (\texttt{M}) by highway driving, and low stress (\texttt{L}) is considered for rest periods. 
For hcilab and  AffectiveROAD databases, the labels were built based on the provided metric. The physiological signals were synchronized with the subjective scores proposed by the latter two databases. A min-max normalization was applied to the scores (workload score in the case of hciLab database and stress score for AffectiveROAD). The final score corresponds to the average of the normalized score values of each extracted segment. 
The range is equally subdivided by three and for each range the label is affected. If the score is between $0$ and $0.33$, the class low (\texttt{L}) is affected to the segment while high (labelled \texttt{H}) is considered when the score is between $0.67$ and $1$. This metric subdivision is made in order to obtain labels consistent with the stress labels extracted from the drivedb database.   

\subsection{Drive profiling and task assignment with spectral clustering} \label{sec:spectralclustering}

All three datasets described in Sec.\ref{sec:data} contain data from multiple drives. Following \cite{dlmNIPS2017,dlmICPR2018}, we group $E$ drives into different \textit{profiles} based on the unique physiological responses of the drivers corresponding to driving-induced affective state, for each drive $d$ in the dataset. These profiles are then used to define the tasks of our multi-view multi-task machine learning model (see Sec.\ref{sec:ml}), such that each task in the model corresponds to a distinct profile.

To build the profiles, we first obtain drive descriptors. This is done by computing the mean value of each feature for all instances with label \texttt{H} in the training set, for each drive. Hence, each drive $d$ is represented by a $D$-dimensional profile vector $\bm{p}_d = [p_{d,1}, ..., p_{d,D}]$, where $D$ is the number of features and $p_{d,i} = {({1}/{N_{tr}^{(d)}})} \sum_j x_i^{(d,j)}$, such that $\bm{x}^{(d,j)}$ is the $j$-th feature vector corresponding to drive $d$ in the training set,  $ x_i^{(d,j)}$ is the $i$-th feature in such vector, and $N_{tr}^{(d)}$ is the number of training instances for drive $d$.

Drive descriptors are then used to cluster drives into $T$ different \textit{profiles}. To this end, we used normalized spectral clustering (see Alg.\ref{alg:spectralclustering}) \cite{ShiMalik2000}. First, we construct a fully connected similarity graph and the corresponding weighted adjacency matrix $W$ using the radial basis function (RBF) kernel for the edge weights, such that $w_{i,j} = K(\bm{p}_i, \bm{p}_j) = \exp{ (-\gamma || \bm{p}_i - \bm{p}_j ||^2) }$ with $\gamma=0.1$. Then, we build the degree matrix $\bm{G}$ as the diagonal matrix with degrees $g_1,...,g_E$ on the diagonal, where $g_i = \sum_{j=1}^{E} w_{ij}$ and $E$ is the number of drives in the dataset. Following this, we compute the unnormalized graph Laplacian $\bm{L}= \bm{G}-\bm{W}$, and derive the first $T$ eigenvectors $\bm{u}_1,...,\bm{u}_T$, where $T$ is the desired number of clusters or profiles. Let $\bm{U} \in \mathbb{R}^{E \times T}$ be the matrix containing the vectors $\bm{u}_1, ..., \bm{u}_T$ as columns, and $\bm{v}_i\in \mathbb{R}^T$ be the vector corresponding to the $i$-th row of $\bm{U}$. We cluster the points $(\bm{v}_i)_{i=1,...,E}$ using the k-means clustering algorithm and produce clusters $C_1, ..., C_T$.

Therefore, each drive $d_1, ..., d_E$ is assigned to one of the $T$ profiles $C_1, ..., C_T$, containing drives that share similar physiological responses to driving-induced affective states. During testing, this clustering process requires access to training data corresponding to the same drive, so that the drive may be assigned to one of the pre-specified profiles. 
The desired number of clusters or profiles is determined by visual inspection of the grouped elements of $\bm{W}$ (see Fig.\ref{fig:clusters}).

\begin{algorithm}
\caption{Normalized spectral clustering \cite{ShiMalik2000}}\label{alg:spectralclustering}
\begin{algorithmic}[1]
\Require Similarity matrix $\bm{S} \in \mathbb{R}^{E\times E}$, number $T$ of clusters to construct.
\State Construct the weighted adjacency matrix $\bm{W}$ of the  similarity graph.
\State Compute the unnormalized Laplacian $\bm{L}$.
\State Compute the first $T$ generalized eigenvectors $\bm{u}_1, ..., \bm{u}_T$ of the generalized eigenproblem $\bm{L} \bm{u}=\lambda \bm{G} \bm{u}$, where $\bm{G}$ is a diagonal matrix.
\State Let $\bm{U} \in \mathbb{R}^{E \times T}$ be the matrix containing the vectors $\bm{u}_1, ..., \bm{u}_T$ as columns.
\State For $i=1, ..., E$, let $\bm{v}_i \in \mathbb{R}^T$ be the vector corresponding to the $i$-th row of $\bm{U}$.
\State Cluster the points $(\bm{v}_i)_{i=1,...,E}$ in $\mathbb{R}^T$ with the k-means algorithm into clusters $C_1, ..., C_T$. \\
\Return Clusters $P_1, ..., P_T$ with $P_i = \{ j | \bm{v}_i \in C_i \}$
\end{algorithmic}
\end{algorithm}

\subsection{Personalized machine learning} \label{sec:ml}

In this work, we consider the following binary classification problem. We have $N$ instances $z_i=(\bm{x}_i,y_i, d_i)$, $i=1,..., N$, where $\bm{x}_i \in \mathbb{R}^D$  is a $D$-dimensional feature vector containing the normalized skin conductance and heart rate features described in Sec.\ref{sec:featureslabels}, $y_i \in \{-1,1 \}$ is a binary label corresponding to \texttt{L} and \texttt{H} instances respectively, and $d_i$ is the drive. Our goal is to learn a decision function $f$ that infers $y$ given $x$ in unseen data. 

\subsubsection{Single-task models}
First, we consider the traditional single-task supervised learning scenario in which the models are drive-agnostic. The goal is to learn a decision function $f$, such that $f(\bm{x_i}) \approx y_i$, using the training instances $\mathcal{D}_{tr} = \{ (\bm{x}_i, y_i )\}_{i=1}^{N_{tr}}$, where $N_{tr}$ is the number of instances in the training set $\mathcal{D}_{tr}$. Specifically, we consider two common machine learning models for binary classification: logistic regression (LR) and support vector machine (SVM).

\subsubsection{Multi-view multi-task model} This learning scenario combines two concepts: (a) multi-view learning and (b) multi-task learning. Views refer to the different signal modalities (EDA or HR) and corresponding features, whereas tasks correspond to the profiles described in Sec.\ref{sec:spectralclustering}. Specifically, we consider 2 views and $T=\{1,2,3\}$ tasks, corresponding to the clusters defined in Sec.\ref{sec:spectralclustering}.

Multi-view learning (MVL) enables combining multiple sensors or signal modalities (views) while automatically learning the importance of each view. In this paper, we consider $M=2$ views (EDA and HR signals) and use a specific multi-view learning technique called multiple-kernel learning (MKL), which allows using multiple views in a kernel-based  algorithm while automatically revealing which views are most useful for solving the classification task. Each signal modality or view $m$ is represented by one kernel $k_m$, and these are subsequently combined into a signal kernel $k_\eta$ by a function $f_\eta$ such that $k_\eta (\bm{x}_i,\bm{x}_j; \bm{\eta}) = f_\eta( \{ k_m (\bm{x}^{(m)}_i,\bm{x}^{(m)}_j) \}_{m=1}^{M} ; \bm{\eta})$, where $\bm{\eta}$ is a parametrization vector that is learned from the data. In this work, we define $f_\eta$ as the weighted average of the kernels, with nonnegative weights that sum up to one:

\begin{equation}
k_\eta (\bm{x}_i,\bm{x}_j; \bm{\eta}) = \sum_{m=1}^M \eta_m k_m (\bm{x}^{(m)}_i,\bm{x}^{(m)}_j)
\end{equation}

The other concept, multi-task learning (MTL), focuses on learning several related prediction tasks simultaneously using a shared representation \cite{Caruana1997}, where each task corresponds to each of the different profiles defined in Sec.\ref{sec:spectralclustering}. 
Hence, in MTL we have a training set $\mathcal{D}^{(r)} = \{ (\bm{x}^{(r)}_i, y^{(r)}_i )\}_{i=1}^{N^{(r)}_{tr}}$ for each profile $r$, and the goal is to train $T$ decision functions $f_r$, one for each profile. 
Because in MTL expertise transfers between tasks, MTL is able to exploit the limited amount of training data available for each task, to the benefit of all.

Multi-task multiple kernel learning (MT-MKL) \cite{Kandemir2014}, combines both concepts: (1) MVL to account for different signal modalities, and (2) MTL to personalize the models according to the different drive profiles. Following \cite{Kandemir2014}, model parameters are learned by solving the following min-max optimization problem:

{ \footnotesize
\begin{equation} \label{eq:optimzation}
\underset{ \left \{ \bm{\eta}^{(r)} \in \mathcal{E} \right \}_{r=1}^{T}  }{\text{minimize}}
\underbrace{
\left \{ 
\underset{ \left \{ \bm{\alpha}^{(r)} \in \mathcal{A}^{(r)} \right \}_{r=1}^{T} }{\text{maximize}}  \Omega ( \{ \bm{\eta}^{(r)} \}_{r=1}^{T} )
+ \sum_{r=1}^T J^{(r)} (\bm{\alpha}^{(r)},\bm{\eta}^{(r)} )
\right \}
}_{\mathcal{O}_\eta}
\end{equation}
}
where $\Omega (\cdot )$ is the regularization term that imposes similarity between the kernels, $\mathcal{E} = \{ \bm{\eta} :\sum_{m=1}^{M} \eta_m = 1, \eta_m \geq 0 \; \forall m \}$ denotes the domain of the kernel combination parameters $\bm{\eta}^{(r)}$, $\mathcal{A}^{(r)}$ is the domain of the parameters of the kernel-based learner (the Lagrange multipliers) for task $r$, and $J^{(r)} (\cdot,\cdot)$ is the objective function of the kernel-based learner of task $r$.

Following \cite{Kandemir2014,dlmICPR2018}, we consider two types of regularizers $\Omega (\cdot)$, the $\ell_1$-norm  and $\ell_2$-norm regularizers respectively:
\begin{equation}
\Omega_1 ( \{ \bm{\eta}^{(r)} \}_{r=1}^T ) = -\nu \sum_{r=1}^T \sum_{s=1}^T  {\bm{\eta}^{(r)}}^\top \bm{\eta}^{(s)}
\end{equation}
\begin{equation}
\Omega_2 ( \{ \bm{\eta}^{(r)} \}_{r=1}^T ) = -\nu \sum_{r=1}^T \sum_{s=1}^T  || \bm{\eta}^{(r)} - \bm{\eta}^{(s)}   ||_2
\end{equation}
where the coefficient $\nu$ controls the influence of the regularizer on Eq. \ref{eq:optimzation}, with larger values of  $\nu$ enforcing similar kernel combination parameters across the tasks $r$, which correspond to each of the clusters in Sec.\ref{sec:spectralclustering}.

As in \cite{dlmICPR2018}, we solve the optimization problem in Eq.\ref{eq:optimzation} using a two-step iterative gradient descent algorithm algorithm (see Alg.\ref{alg:mtmkl}), where the gradient of the objective function $\mathcal{O}_n$ is given by:

{ \small
\begin{equation}
\frac{\partial \mathcal{O}_n}{\partial \eta_m^{(r)}} = 
-2 \frac{\partial \bm{\Omega}(\bm{\eta}^{(r)})}{\partial \eta_m^{(r)}}
-\frac{1}{2} \sum_{i=1}^N \sum_{j=1}^N
\alpha_i^{(r)} \alpha_j^{(r)} y^{(r)} y^{(r)} k_m^{(r)} (\bm{x}_i^{(m)},\bm{x}_j^{(m)})
\end{equation}
}

For each task-specific model, we use a least-squares support vector machine (LSSVM) and consider two kernels: (a) the linear kernel $k(\bm{x}, \bm{x}') = \bm{x}^\top \bm{x}'$, and (b) the radial basis function (RBF) kernel $k(\bm{x}, \bm{x}') = \exp (- \gamma || \bm{x}-\bm{x}' ||^2)$.

\begin{algorithm}
\caption{Multitask Multiple Kernel Learning (MT-MKL)}\label{alg:mtmkl}
\begin{algorithmic}[1]
\State Initialize $\bm{\eta}^{(r)}$ as $(1/T, ..., 1/T)$, $\forall r$
\Repeat 
\State Calculate $\bm{K}^{(r)}_\eta = \{ k_\eta^{(r)}  (\bm{x}_i^{(r)}, \bm{x}_j^{(r)}; \bm{\eta}^{(r)}) \}^{N^{(r)}}_{i,j=1} $, $\forall r$
\State Solve a single-kernel machine using $\bm{K}^{(r)}_\eta$, $\forall r$
\State Update $\bm{\eta}^{(r)}$ in the opposite direction of $\partial \bm{\mathcal{O}}_n / \partial \bm{\eta}^{(r)}$, $\forall r$
\Until{convergence}
\end{algorithmic}
\end{algorithm}

\begin{figure}
	\centering
	\includegraphics[width=0.98\linewidth]{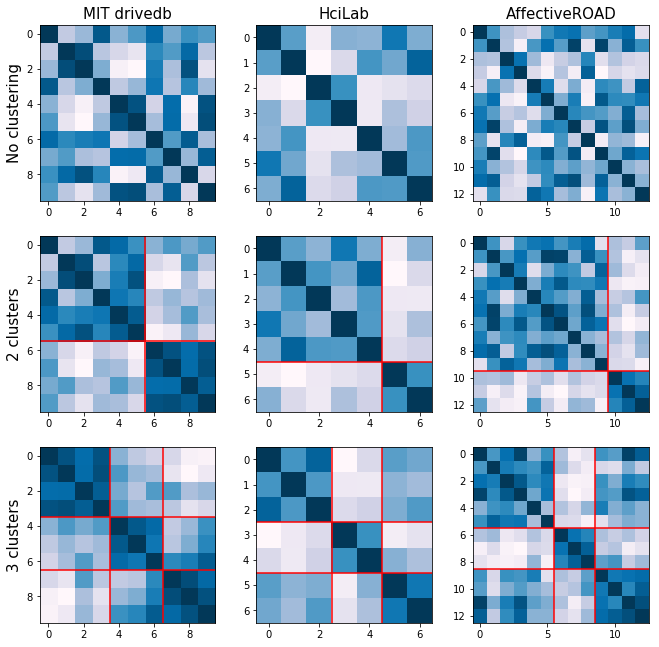}
	\caption{Drive profiling and task assignment using the spectral clustering algorithm. The elements in the  similarity matrices for all three databases and different numbers of clusters $T$ represent the similarity  between any two drives, with darker matrix elements indicating increased similarity. The resulting clusters are indicated in red.}
	\label{fig:clusters}
\end{figure}

\begin{table}
\centering
\caption{Single-task models evaluated using 10-fold cross-validation.}
\label{tab:stml}
\begin{tabular}{lrrrr}
\hline
\textbf{Dataset \& model} & \multicolumn{1}{c}{\textbf{Acc.}} & \multicolumn{1}{c}{\textbf{Pr.}} & \multicolumn{1}{c}{\textbf{R.}} & \multicolumn{1}{c}{\textbf{F$_1$}} \\ \hline
\textbf{A. MIT drivedb}   & \multicolumn{1}{l}{}                  & \multicolumn{1}{l}{}              & \multicolumn{1}{l}{}             & \multicolumn{1}{l}{}               \\
Logistic regression (L1)  & 0.84                                  & 0.66                              & 0.68                             & 0.67                               \\
Logistic regression (L2)  & 0.83                                  & 0.82                              & 0.84                             & 0.83                               \\
SVM (linear kernel)       & \textbf{0.85}                                  & 0.82                              & 0.88                             & 0.85                               \\
SVM (rbf kernel)          & \textbf{0.85}                                  & 0.82                              & 0.90                             & 0.86                               \\
\textbf{B. HciLab}        & \multicolumn{1}{l}{}                  & \multicolumn{1}{l}{}              & \multicolumn{1}{l}{}             & \multicolumn{1}{l}{}               \\
Logistic regression (L1)  & 0.60                                  & 0.63                              & 0.56                             & 0.57                               \\
Logistic regression (L2)  & 0.62                                  & 0.63                              & 0.58                             & 0.58                               \\
SVM (linear kernel)       & \textbf{0.64}                                  & 0.62                              & 0.55                             & 0.58                               \\
SVM (rbf kernel)          & 0.63                                  & 0.64                              & 0.59                             & 0.58                               \\
\textbf{C. AffectiveROAD} & \multicolumn{1}{l}{}                  & \multicolumn{1}{l}{}              & \multicolumn{1}{l}{}             & \multicolumn{1}{l}{}               \\
Logistic regression       & 0.66                                  & 0.66                              & 0.68                             & 0.67                               \\
Logistic regression (L2)  & 0.66                                  & 0.65                              & 0.67                             & 0.66                               \\
SVM (linear kernel)       & 0.67                                  & 0.65                              & 0.75                             & 0.70                               \\
SVM (rbf kernel)          & \textbf{0.70}                                  & 0.67                              & 0.79                             & 0.72                              
\end{tabular}
\end{table}

\begin{table*}
\centering
\caption{Average accuracy (ACC), precision (Pr.), recall (R.) and F$_1$ score of the MT-MKL models over 10-fold cross-validation}
\label{tab:mtmkl}
\small
\begin{tabular}{llllllllllllllllll}
\hline
\multicolumn{3}{l}{\textbf{MT-MKL model}} &  & \multicolumn{4}{l}{\textbf{A. MIT drivedb}}                                                  &                      & \multicolumn{4}{l}{\textbf{B. HciLab}}                                                   &                      & \multicolumn{4}{l}{\textbf{C. AffectiveROAD}}                                                           \\ \cline{1-3} \cline{5-8} \cline{10-13} \cline{15-18} 
T                     & Kernel    & Reg   &  & \multicolumn{1}{c}{ACC} & \multicolumn{1}{c}{Pr.} & \multicolumn{1}{c}{R.} & \multicolumn{1}{c}{F1} & \multicolumn{1}{c}{} & \multicolumn{1}{c}{ACC} & \multicolumn{1}{c}{Pr.} & \multicolumn{1}{c}{R.} & \multicolumn{1}{c}{F1} & \multicolumn{1}{c}{} & \multicolumn{1}{c}{ACC} & \multicolumn{1}{c}{Pr.} & \multicolumn{1}{c}{R.} & \multicolumn{1}{c}{F1} \\ \hline
\multirow{4}{*}{1}    & Linear    & L1    &  & 0.84                    & 0.86                    & 0.86                   & 0.86                   &                      & 0.65                    & 0.40                    & 0.44                   & 0.42                   &                      & 0.62                    & 0.82                    & 0.80                   & 0.81                   \\
                      & Linear    & L2    &  & 0.84                    & 0.86                    & 0.86                   & 0.87                   &                      & 0.64                    & 0.36                    & 0.38                   & 0.37                   &                      & 0.62                    & 0.83                    & 0.81                   & 0.82                   \\
                      & RBF       & L1    &  & 0.91                    & 0.91                    & 0.91                   & 0.91                   &                      & 0.65                    & 0.39                    & 0.40                   & 0.40                   &                      & 0.74                    & 0.81                    & 0.81                   & 0.81                   \\
                      & RBF       & L2    &  & 0.90                    & 0.90                    & 0.90                   & 0.90                   &                      & 0.64                    & 0.36                    & 0.37                   & 0.35                   &                      & 0.73                    & 0.80                    & 0.80                   & 0.80                   \\ \hline
\multirow{4}{*}{2}    & Linear    & L1    &  & 0.85                    & 0.88                    & 0.88                   & 0.88                   &                      & 0.68                    & 0.40                    & 0.42                   & 0.41                   &                      & 0.70                    & 0.84                    & 0.83                   & 0.83                   \\
                      & Linear    & L2    &  & 0.86                    & 0.88                    & 0.88                   & 0.89                   &                      & 0.65                    & 0.39                    & 0.42                   & 0.41                   &                      & 0.70                    & 0.84                    & 0.83                   & 0.83                   \\
                      & RBF       & L1    &  & 0.91                    & 0.91                    & 0.91                   & 0.91                   &                      & 0.67                    & 0.40                    & 0.42                   & 0.40                   &                      & 0.82                    & 0.86                    & 0.86                   & 0.86                   \\
                      & RBF       & L2    &  & 0.91                    & 0.91                    & 0.91                   & 0.91                   &                      & \textbf{0.71}                    & 0.46                    & 0.48                   & 0.48                   &                      & 0.82                    & 0.86                    & 0.87                   & 0.86                   \\ \hline
\multirow{4}{*}{3}    & Linear    & L1    &  & 0.88                    & 0.90                    & 0.90                   & 0.90                   &                      & 0.68                    & 0.41                    & 0.42                   & 0.42                   &                      & 0.69                    & 0.85                    & 0.83                   & 0.84                   \\
                      & Linear    & L2    &  & 0.87                    & 0.90                    & 0.89                   & 0.89                   &                      & 0.68                    & 0.43                    & 0.45                   & 0.42                   &                      & 0.70                    & 0.86                    & 0.85                   & 0.85                   \\
                      & RBF       & L1    &  & \textbf{0.93}                    & 0.93                    & 0.94                   & 0.94                   &                      & 0.69                    & 0.42                    & 0.43                   & 0.41                   &                      & 0.83                    & 0.87                    & 0.87                   & 0.87                   \\
                      & RBF       & L2    &  & 0.93                    & 0.93                    & 0.93                   & 0.93                   &                      & 0.70                    & 0.44                    & 0.47                   & 0.46                   &                      & \textbf{0.83}                    & 0.87                    & 0.87                   & 0.87                   \\ \cline{2-18} 
\end{tabular}
\end{table*}

\section{Results}
Following the feature and label extraction process described in Sec. \ref{sec:featureslabels} for all three datasets, we balanced these to have equal amounts of \texttt{L} and \texttt{H} instances by downsampling the overrepresented class. After this, datasets A, B, C contained $2102$, $194$ and $1504$ samples respectively, with each dataset containing half "high stress" (\texttt{H}) and half "low stress" (\texttt{L}) windows of data from real-world driving, with all features in the normalized range $[0,1]$.

First, we trained standard single-task algorithms using 10-fold cross-validation on the balanced datasets. 
The results are  summarized in  Table \ref{tab:stml}, and show better performance of the SVM model over the LR one, for all three datasets. The maximmum accuracies were $0.85$, $0.64$ and $0.70$ for the MIT drivedb, HciLab and AffectiveROAD datasets respectively.

Following this, we investigated the effect of accounting for drive-specific differences in physiological responses. As described in Sec.\ref{sec:spectralclustering}, for each drive $d$ we calculated a drive descriptor vector $\bm{p}_d$ using  training data only. These descriptor vectors were then used to assign each drive to a profile or cluster (with $T=\{1,2,3\}$ clusters) using spectral clustering. The results of the clustering process are shown in Fig. \ref{fig:clusters}. Each of these clusters defined a drive \textit{profile}, which represented a task in our MT-MKL algorithm. As before, to train and evaluate the MT-MKL model, we performed 10-fold cross-validation. To optimize the model hyperparameters, 5-fold cross-validation was performed within the training set. Specifically, we optimized $C$ and $\nu$, which were selected from the set $\{ 10^{-4}, ..., 10^{2} \}$. In the models with radial basis function (RBF) kernel, we also optimized $\gamma$ by selecting the best performing value from $\{ 10^{-1}, ...,10\}$ when evaluated in the validation set, which was comprised of 10\% of the data not present in the test set.

The results of the multi-view multi-task model for all three datasets are shown in Table. \ref{tab:mtmkl}, and indicate an overall improvement in performance as compared with the single-task models, even when no clustering is performed. They also show improved performance as we increase the number of clusters. For the MIT drivedb and AffectiveROAD databases, the best performances were obtained when 3 clusters are used, resulting in $93\%$ and $83\%$ classification accuracy respectively. For the HciLab, the best performance ($71\%$) was obtained with 2 clusters. This suggests that the drives can be grouped into three driving profiles for drivedb and AffectiveROAD databases, whereas only two driving groups seem relevant to consider in the case of the HciLab database. This similarity between drivedb and AffectiveROAD in the clustering results may be attributed to the fact that the drives were performed based on the same driving protocol. Furthermore, we hypothesize that the small number of drives in this dataset results in no additional benefit of an increased number of clusters. In all cases, best performances were obtained with the RBF kernel.

\begin{figure}
	\centering
	\includegraphics[width=0.59\linewidth]{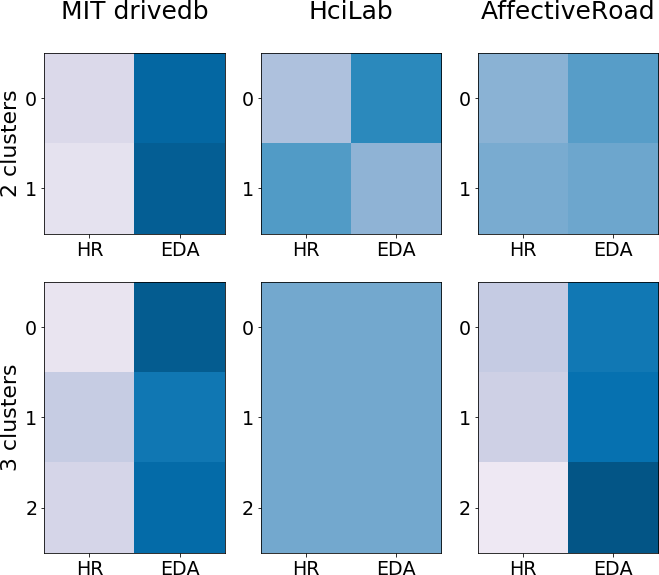}
	\caption{Kernel weights $\eta$ of the multi-view multi-task model. The larger the  weights are, the  darker  the  matrix  elements  are. This represents  increasing  importance  of that view  for binary classification performance for a given task.}
	\label{fig:kernelweights}
\end{figure}

Finally, we examined the coefficients $\eta \in [0,1]$ corresponding to the best performing models for $1$, $2$, $3$ clusters and the three databases. These are depicted in Fig.\ref{fig:kernelweights} and show overall larger coefficients for the EDA view, suggesting an increased importance of this signal in the binary classification task. This result confirms the finding of \cite{ElHaouijSMA} where the EDA was found more important compared to the different physiological signals, especially HR, when using a random forest approach for stress level classification on the  MIT drivedb database.

\section{Conclusion}
In this work, we present an approach for assessment of car driver's affective state in the context of real-world driving, using personalized machine learning. Specifically, we employed a multi-view multi-task machine learning framework that is able to account for inter-subject inter-drive variability in affective responses to the driving experience, while providing interpretability of the relative importance of the different sensor modalities in  model performance. 
We tested our approach in three different databases containing physiological data from real-world driving experiences, resulting in $93\%$, $71\%$ and $83\%$ classification accuracies respectively. The differences in model performance may be explained by inconsistent driving protocols and labels, different database sizes or the use of different sensors
for data collection. Our  results also indicate that accounting for drive-specific differences significantly improves model  performance, and that the electrodermal activity signals tend to play a more important role than the heart rate data. 

This work has several limitations that should be addressed in future work. Whereas this work focused exclusively on classifying \texttt{L} versus \texttt{H}, the intermediate stress level \texttt{M} was not evaluated. Furthermore, the proposed driving clusters (Fig.\ref{fig:clusters}) should be investigated in terms of driving conditions, driving styles and driver demographics. 
Future work should also include other signal modalities (e.g. facial video, breathing rate, etc.), more granular affective states and different multi-task machine learning methods (e.g. multi-task neural networks \cite{Saeed2017,dlmACII2017}).

\bibliographystyle{ieeetr}
\bibliography{Bib,Bib_Daniel}

\begin{thebibliography}{10}

\bibitem{Nass2005}
C.~Nass, I.-M. Jonsson, H.~Harris, B.~Reaves, J.~Endo, S.~Brave, and
  L.~Takayama, ``{Improving automotive safety by pairing driver emotion and car
  voice emotion},'' in {\em ACM Conference on Human Factors in Computing
  Systems (CHI)}, (New York, New York, USA), p.~1973, ACM Press, 2005.

\bibitem{Johnsson2005}
I.-M. Johnsson, C.~Nass, H.~Harris, and L.~Takayama, ``{Matching In-Car Voice
  with Driver State : Impact on Attitude and Driving Performance},'' in {\em
  Driving assessment 2005: proceedings of the 3rd International Driving
  Symposium on Human Factors in Driver Assessment, Training, and Vehicle
  Design}, (Iowa City, Iowa), pp.~173--180, University of Iowa, 10 2005.

\bibitem{Healey2005}
J.~Healey and R.~Picard, ``Detecting stress during real-world driving tasks
  using physiological sensors,'' {\em IEEE Transactions on Intelligent
  Transportation Systems}, vol.~6, pp.~156--166, June 2005.

\bibitem{Rigas2008}
G.~Rigas, C.~D. Katsis, P.~Bougia, and D.~I. Fotiadis, ``{A reasoning-based
  framework for car driver'ss stress prediction},'' in {\em 2008 16th
  Mediterranean Conference on Control and Automation}, pp.~627--632, IEEE, 6
  2008.

\bibitem{Lee2012}
B.~{Lee} and W.~{Chung}, ``Driver alertness monitoring using fusion of facial
  features and bio-signals,'' {\em IEEE Sensors Journal}, vol.~12,
  pp.~2416--2422, July 2012.

\bibitem{Singh2013}
R.~R. Singh, S.~Conjeti, and R.~Banerjee, ``{A comparative evaluation of neural
  network classifiers for stress level analysis of automotive drivers using
  physiological signals},'' {\em Biomedical Signal Processing and Control},
  vol.~8, pp.~740--754, nov 2013.

\bibitem{Vincente2016}
J.~Vicente, P.~Laguna, A.~Bartra, and R.~Bail{\'{o}}n, ``{Drowsiness detection
  using heart rate variability},'' {\em Medical {\&} Biological Engineering
  {\&} Computing}, vol.~54, pp.~927--937, jun 2016.

\bibitem{Chen2017}
L.-l. Chen, Y.~Zhao, P.-f. Ye, J.~Zhang, and J.-z. Zou, ``{Detecting driving
  stress in physiological signals based on multimodal feature analysis and
  kernel classifiers},'' {\em Expert Systems with Applications}, vol.~85,
  pp.~279--291, nov 2017.

\bibitem{Saeed2017}
A.~Saeed and S.~Trajanovski, ``Personalized driver stress detection with
  multi-task neural networks using physiological signals,'' in {\em Neural
  Information Processing Systems (NIPS) Workshop on Machine Learning for
  Health}, (Long Beach, CA, USA), December 2017.

\bibitem{Kim2018}
S.~Kim, W.~Rhee, D.~Choi, Y.~J. Jang, and Y.~Yoon, ``{Characterizing Driver
  Stress Using Physiological and Operational Data from Real-World Electric
  Vehicle Driving Experiment},'' {\em International Journal of Automotive
  Technology}, vol.~19, pp.~895--906, oct 2018.

\bibitem{ElHaouijSMA}
N.~El~Haouij, J.-M. Poggi, R.~Ghozi, S.~Sevestre-Ghalila, and M.~{Ja{\"i}dane},
  ``Random forest-based approach for physiological functional variable
  selection for driver's stress level classification,'' {\em Statistical
  Methods {\&} Applications}, Feb 2018.

\bibitem{doshiKim2017}
F.~Doshi-Velez and B.~Kim, ``{Towards A Rigorous Science of Interpretable
  Machine Learning},'' in {\em eprint arXiv:1702.08608}, 2 2017.

\bibitem{Goldberger2000}
A.~L. Goldberger, L.~A.~N. Amaral, L.~Glass, J.~M. Hausdorff, P.~C. Ivanov,
  R.~G. Mark, J.~E. Mietus, G.~B. Moody, C.-K. Peng, and H.~E. Stanley,
  ``{PhysioBank, PhysioToolkit, and PhysioNet},'' {\em Circulation}, vol.~101,
  6 2000.

\bibitem{Databases}
``{MIT Media Lab, Affective Computing Group databases}.''
  \url{https://affect.media.mit.edu/share-data.php}.

\bibitem{Schneegass2013}
S.~Schneegass, B.~Pfleging, N.~Broy, A.~Schmidt, and H.~F., ``{ A data set of
  real world driving to assess driver workload},'' in {\em 5th International
  Conference on Automotive User Interfaces and Interactive Vehicular
  Applications (AutomotiveUI'13). ACM, New York, NY, USA}, pp.~150--157, IEEE,
  sep 2013.

\bibitem{ElHaouij2018}
N.~El~Haouij, J.-M. Poggi, S.~Sevestre-Ghalila, R.~Ghozi, and M.~{Ja{\"i}dane},
  ``{AffectiveROAD System and Database to Assess Driver's Arousal State},'' in
  {\em SAC 2018: Symposium on Applied Computing , \textsl{April} 9--13, 2018,
  Pau, France}, 2018.

\bibitem{Boucsein2012}
W.~Boucsein, {\em {Electrodermal Activity}}.
\newblock Boston, MA: Springer US, 2012.

\bibitem{Healey2000}
J.~Healey, {\em Wearable and automotive systems for affect recognition from
  physiology}.
\newblock PhD thesis, MIT Dept. of Electrical Engineering and Computer Science,
  2000.

\bibitem{dlmNIPS2017}
D.~Lopez-Martinez, O.~Rudovic, and R.~Picard, ``{Physiological and Behavioral
  Profiling for Nociceptive Pain Estimation Using Personalized Multitask
  Learning},'' in {\em Neural Information Processing Systems (NIPS) Workshop on
  Machine Learning for Health}, (Long Beach, USA), 2017.

\bibitem{dlmICPR2018}
D.~Lopez-Martinez, K.~Peng, S.~Steele, A.~Lee, D.~Borsook, and R.~Picard,
  ``{Multitask multiple kernel machines for personalized pain recognition from
  functional near-infrared spectroscopy brain signals},'' in {\em International
  Conference on Pattern Recognition (ICPR)}, (Beijing), 2018.

\bibitem{ShiMalik2000}
{Jianbo Shi} and J.~Malik, ``{Normalized cuts and image segmentation},'' {\em
  IEEE Transactions on Pattern Analysis and Machine Intelligence}, vol.~22,
  no.~8, pp.~888--905, 2000.

\bibitem{Caruana1997}
R.~Caruana, ``{Multitask Learning},'' {\em Machine Learning}, vol.~28, no.~1,
  pp.~41--75, 1997.

\bibitem{Kandemir2014}
M.~Kandemir, A.~Vetek, M.~G{\"{o}}nen, A.~Klami, and S.~Kaski, ``{Multi-task
  and multi-view learning of user state},'' {\em Neurocomputing}, vol.~139,
  pp.~97--106, 9 2014.

\bibitem{dlmACII2017}
D.~Lopez-Martinez and R.~Picard, ``{Multi-task neural networks for personalized
  pain recognition from physiological signals},'' in {\em 2017 Seventh
  International Conference on Affective Computing and Intelligent Interaction
  Workshops and Demos (ACIIW)}, pp.~181--184, IEEE, 10 2017.

\end{thebibliography}

\end{document}